\documentclass[conference]{IEEEtran}
\IEEEoverridecommandlockouts
\usepackage{cite}
\usepackage{amsmath,amssymb,amsfonts}
\usepackage{algorithmic}
\usepackage{graphicx}
\usepackage{textcomp}
\usepackage{xcolor}

\usepackage{hyperref}
\usepackage{graphicx}
\usepackage[inkscapeformat=png]{svg}
\usepackage{booktabs}

\usepackage{braket}
\usepackage{flushend}

\def\BibTeX{{\rm B\kern-.05em{\sc i\kern-.025em b}\kern-.08em
    T\kern-.1667em\lower.7ex\hbox{E}\kern-.125emX}}
\begin{document}

\title{Q-SCALE: Quantum computing-based Sensor Calibration for Advanced Learning and Efficiency
}

\author{
Lorenzo Bergadano$^{1}$,
Andrea Ceschini$^{2}$,
Pietro Chiavassa$^{1}$,
Edoardo Giusto$^{3*}$,\\
Bartolomeo Montrucchio$^1$
Massimo Panella$^{2}$,
Antonello Rosato$^{2}$,\\
$^1$ Polytechnic of Turin, Italy\\
$^2$ University of Rome ``La Sapienza'', Italy\\
$^3$ University of Naples, Federico II, Italy $^*$egiusto@ieee.org\\
}

\maketitle

\thispagestyle{plain}
\pagestyle{plain}

\begin{abstract}
In a world burdened by air pollution, the integration of state-of-the-art sensor calibration techniques utilizing Quantum Computing (QC) and Machine Learning (ML) holds promise for enhancing the accuracy and efficiency of air quality monitoring systems in smart cities.
This article investigates the process of calibrating inexpensive optical fine-dust sensors through advanced methodologies such as Deep Learning (DL) and Quantum Machine Learning (QML). The objective of the project is to compare four sophisticated algorithms from both the classical and quantum realms to discern their disparities and explore possible alternative approaches to improve the precision and dependability of particulate matter measurements in urban air quality surveillance.
Classical Feed-Forward Neural Networks (FFNN) and Long Short-Term Memory (LSTM) models are evaluated against their quantum counterparts: Variational Quantum Regressors (VQR) and Quantum LSTM (QLSTM) circuits.
Through meticulous testing, including hyperparameter optimization and cross-validation, the study assesses the potential of quantum models to refine calibration performance.
Our analysis shows that: the FFNN model achieved superior calibration accuracy on the test set compared to the VQR model in terms of lower L1 loss function ($2.92$ vs $4.81$); the QLSTM slightly outperformed the LSTM model (loss on the test set: $2.70$ vs $2.77$), despite using fewer trainable weights ($66$ vs $482$).

\end{abstract}


\section{Introduction}
In the quest for smarter, more sustainable urban environments, the precise monitoring of air quality stands out as a cornerstone for public health and environmental policy \cite{IDREES2020100123}.
In this domain, the calibration of optical fine dust sensors 
emerges as a critical challenge, especially when the focus shifts toward low-cost, scalable solutions that can be deployed extensively across urban landscapes.
This scientific inquiry delves into the application of advanced computational models, particularly Deep Learning (DL) and Quantum Machine Learning (QML), to calibrate these sensors, thereby enhancing their accuracy and reliability in measuring particulate matter concentrations.

Urban centers worldwide face the challenge of air pollution, a silent yet pervasive threat to health and the environment that can be treated by using suited data-driven modeling approaches based on machine learning and deep learning tools \cite{PANELLA2003521,FENG2015118,CABANEROS2019285,8732985}. 
Fine particulate matter ($\mathrm{PM}_{2.5}$) is among the most hazardous pollutants, with the potential to penetrate deep into the respiratory tract and bloodstream, causing long-term health issues.
The accurate monitoring of these pollutants is not just a technical challenge but a societal imperative, forming the basis for informed policy decisions and public health initiatives \cite{okokpujie2018smart}.

Low-cost optical fine dust sensors offer a promising avenue for widespread air quality monitoring, aligning with the vision of smart cities \cite{9247550,giusto2018etfa}.
However, their utility is hampered by calibration challenges, which stem from the sensors' susceptibility to environmental variables and their limited accuracy compared to high-cost counterparts used by official environmental agencies.
Addressing these limitations requires innovative approaches that can learn from complex datasets and adapt to varying conditions, ensuring the sensors' outputs are both reliable and precise.


The contributions of this paper are the following:
\begin{itemize}
    \item Proposing the application of advanced Machine Learning (ML) techniques like DL and QML for calibrating low-cost optical fine dust sensors to improve air quality monitoring accuracy.
    \item Implementing and comparing four different neural models: classical Feed-Forward Neural Network (FFNN), Long Short-Term Memory (LSTM) network, and their quantum counterparts, i.e. Variational Quantum Regressor (VQR) and Quantum LSTM (QLSTM) for time series forecasting. Each model is carefully selected after an extensive hyperparameter tuning procedure, thus resulting in the best possible model for the underlying task.
    \item Highlighting the potential computational efficiency and scalability advantages of quantum models like QLSTM, which achieved comparable results with only 66 parameters compared to 482 for classical LSTM.
    \item Contributing to the broader research on applying QML to solve complex real-world problems beyond the current theoretical studies.
\end{itemize}
    
The rest of the paper is organized as follows:
Section \ref{sec:related_works} discusses related works in the domain of ML applied to air pollution monitoring applications; Section \ref{sec:background} describes classical and quantum models employed in this work; Section \ref{sec:methodology} explains the adopted approach and the considered data; Section \ref{sec:results} describes and analyzes the results obtained by the different methods. Eventually, Section \ref{sec:conclusions} lays out conclusions and future work perspectives.

\section{Related Works}
\label{sec:related_works}
The state-of-the-art in air pollution monitoring includes different approaches, each with its unique advantages and limitations.

DL models, given their modularity and the ability to learn hierarchical representations of data, have been particularly successful in addressing the nonlinear and complex nature of environmental phenomena \cite{yuan2020deep}. 
For example, LSTM networks have been effectively used to model time-series data from sensors, capturing both short-term fluctuations and long-term trends in air quality indicators \cite{jiao2019prediction}.

Satellite monitoring stands out for its extensive area coverage and ability to access remote locations. 
It provides valuable long-term data, enabling trend analysis and real-time air quality assessment \cite{holloway2021satellite}. 
However, its efficacy is sometimes hindered by small-scale pollution events, adverse weather conditions, and the high costs associated with satellite technology development and maintenance.

Ground-based stations offer precise pollutant measurements and are indispensable for localized data collection, crucial for policy-making. 
Despite their precision, these stations cover limited areas and require a dense network for comprehensive coverage \cite{filonchyk2019urban}. 
Their operation demands regular maintenance and calibration, posing financial and logistical challenges.

The integration of fine-dust sensors into Internet of Things (IoT)~\cite{s151229859} networks is an emerging solution for air quality monitoring within smart cities.
This approach ensures real-time data availability and facilitates extensive data analysis. 
However, the accuracy of such sensors can be influenced by environmental variables, which requires regular calibration \cite{chaturvedi2020iot}.
Good practices for calibration and metrics for data quality assessments are discussed in the literature~\cite{eval_cal_practices}.
Performance evaluations are conducted either in the lab~\cite{eval_lab} or in the field~\cite{eval_how_to_best, eval_long_term, eval_filippo} by comparing the low-cost sensors to reference instruments.
Simulation software has also been developed to study their behavior~\cite{eval_physics}.

However, despite the significant advancements in air pollution monitoring techniques within the DL realm, one notable area that has yet to be explored is the application of quantum models. 
Recently, Quantum Neural Networks (QNNs) have emerged as a promising trend in DL \cite{chen2020novel,ceschini_2021,panella_2009}. Leveraging quantum superposition and entanglement in complex high-dimensional states, QNNs efficiently handle large-scale heterogeneous data and conduct high-dimensional processing with a reduced number of qubits, resulting in faster computation and lower error rates thanks to the Variational Quantum Circuit (VQC) paradigm \cite{tacchinoArtificialNeuronImplemented2019a,abbasPowerQuantumNeural2021}, even on Noisy Intermediate-Scale Quantum (NISQ) devices \cite{markidis2023programming}. 
To our knowledge, no attempt has been made to apply QML to the challenge of air quality monitoring. In this work, we compare different classical and quantum neural models in a time series forecasting problem, aiming to attain more robust results and efficient computation in predicting $\mathrm{PM}_{2.5}$ levels compared to traditional methods.

\section{Background}
\label{sec:background}
We hereby present the architectures and the neural models employed in our paper, i.e. plain VQCs, FFNN, VQR, LSTM and QLSTM. 

\subsection{Variational Quantum Circuits (VQCs)}

Quantum computing utilizes quantum mechanics to fundamentally alter information processing compared to traditional computing \cite{10.1145/367701.367709}. 
It does so by employing qubits, which are capable of handling extensive data via superposition and entanglement, potentially outperforming classical computers in various fields, including cryptography and materials science.
QML combines quantum computing with artificial intelligence to potentially improve ML models, thus seeking quantum advantage \cite{Biamonte2017}. 

VQCs are gaining traction in QML due to their adaptability in solving tasks using classical optimization methods on NISQ devices \cite{Cerezo2021}. 
They offer a practical way to utilize quantum benefits despite noise and limited qubit stability. 
In particular, VQCs use parametrized quantum gates, optimized to minimize a cost function, bridging quantum and classical computing strengths in a hybrid approach.

A VQC is represented as a sequence of unitary operations $U(\theta)$ on a quantum state $\ket{\psi_0}$, where $\theta$ denotes the set of parameters controlling the quantum gates.
The unitary nature of these operations ensures the preservation of the quantum state norm.
The action of a VQC on an initial quantum state can be mathematically expressed as:
\begin{equation}
    \label{eq:vqc-action-initial-qs}
    \ket{\psi(\theta)}=U(\theta)\ket{\psi_0},
\end{equation}
where $\ket{\psi(\theta)}$ symbolizes the parameter-dependent quantum state.
The objective of such variational algorithms is to iteratively adjust $\theta$ to minimize a cost function $C(\theta)$, typically formulated as the expectation value of an observable $O$ with respect to the parameterized state:
\begin{equation}
    \label{eq:vqc-minimize-cost-function}
    C(\theta)=\bra{\psi(\theta)}O\ket{\psi(\theta)}.
\end{equation}

VQCs can outperform traditional methods like classical ML for complex tasks due to their ability to leverage quantum parallelism and enhanced expressive power in a high-dimensional Hilbert space.
Quantum parallelism enables VQCs to explore vast solution spaces simultaneously, potentially leading to faster convergence and more efficient solutions; meanwhile, the flexibility and expressiveness of VQCs allow them to represent complex functions and capture intricate patterns and relationships in data that might be challenging for classical models to learn.
In particular, VQCs are the basis for the quantum models employed in this work, i.e. VQR and QLSTM. The core idea is to leverage hidden correlations among features in time series through superposition and entanglement, thereby improving the predictive capabilities of the quantum models.

\subsection{Feed-Forward Neural Network (FFNN)}
The FFNN model
consists of an input layer, several hidden layers, and an output layer \cite{329294}. 
Each layer comprises neurons that perform weighted sums of their inputs followed by a non-linear activation function. 
The mathematical representation of a neuron's operation in layer $l$ is given by:
\begin{equation}
    \label{eq:single-neuron-operation}
    a^{(l)} = \sigma(W^{(l)}a^{(l-1)}+b^{(l)}),
\end{equation}
where $a^{(l)}$ is the activation of layer $l$, $W^{(l)}$ is the weight matrix, $b^{(l)}$ is the bias vector, and $\sigma$ is the activation function.
The FFNN used in this study was optimized to predict particulate matter values by minimizing the loss function through backpropagation, adjusting $W$ and $b$ to fit the training data. We chose such a simple model to set a standard benchmark for our prediction task. 

\subsection{Long Short-Term Memory (LSTM)}
LSTM networks are designed to address the limitations of traditional RNNs in learning long-term dependencies \cite{SHERSTINSKY2020132306}. 
The core of an LSTM unit includes three gates: the input gate $i$, the forget gate $f$, and the output gate $o$, which regulate the flow of information. 
The cell state $C_t$ acts as the unit's memory.
The operations within an LSTM cell can be summarized as follows:
\begin{itemize}
    \item \textbf{Forget Gate:} 
    \[f_t=\sigma(W_f\cdot[h_{t-1},x_t]+b_f);\]
    \item \textbf{Input Gate:} 
    \[
    \begin{aligned}
    &i_t=\sigma(W_i\cdot[h_{t-1},x_t]+b_i),\\
    &\Tilde{C_t}=\tanh(W_C\cdot[h_{t-1},x_t]+b_C);
    \end{aligned}
    \]
    \item \textbf{Cell State Update:} 
    \[C_t=f_t*C_{t-1}+i_t*\Tilde{C_t};\]
    \item \textbf{Output Gate:} 
    \[
    \begin{aligned}
    &o_t=\sigma(W_o\cdot[h_{t-1},x_t]+b_o),\\
    &h_t=o_t*\tanh(C_t).
    \end{aligned}
    \]
\end{itemize}
This architecture enables LSTM networks to effectively capture temporal patterns in particulate matter concentration data over extended periods, thus it is particularly indicated for time series forecasting.

\subsection{Variational Quantum Regressor (VQR)}
\begin{figure*}[htbp]
    \centerline{\includegraphics[width=\linewidth]{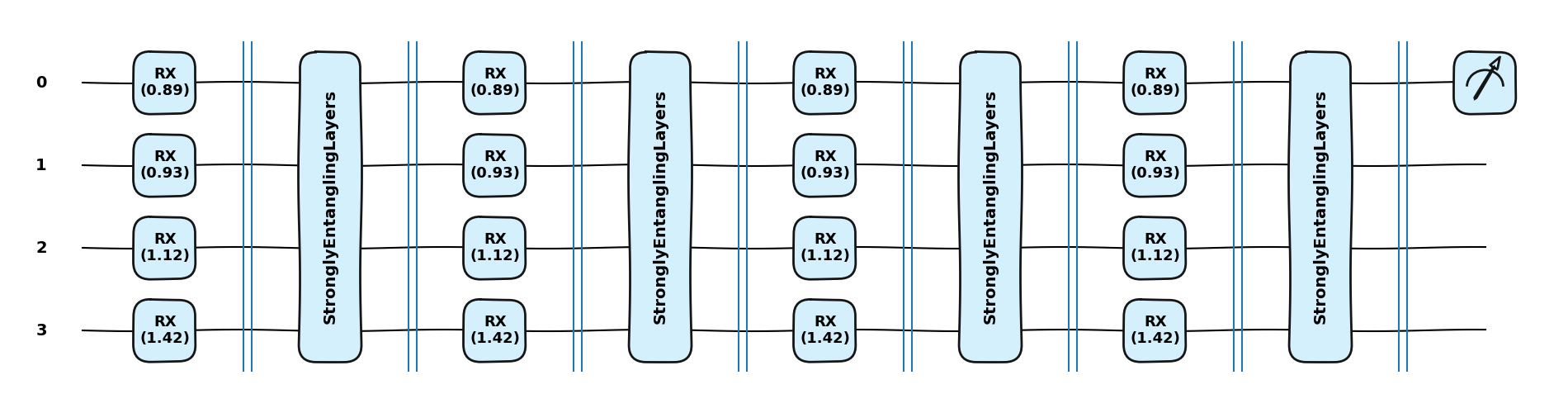}}
    \caption{The Non Linear VQR circuit implementation with 4 qubits, made with the Pennylane library.}
    \label{fig:vqr-non-linear}
\end{figure*}
The VQR model integrates the principles of quantum computing into ML through VQCs.
These circuits use parameterized quantum gates to encode data and learn from it, exploiting the quantum mechanical phenomena of superposition and entanglement to process information.
The VQR's objective is to optimize a set of parameters $\theta$ to minimize a cost function, which in this context is the difference between the predicted and actual $\mathrm{PM}_{2.5}$ values.
The optimization is performed using quantum gradient descent, leveraging the quantum circuit's ability to efficiently compute gradients through the parameter-shift rule~\cite{param_shift_rule}.

The VQR operates by first encoding the input data into the quantum circuit, utilizing the qubits' ability to represent data in a multidimensional complex state space.
The circuit then processes this information through its parameterized quantum gates, effectively performing the regression task by mapping inputs to predicted outputs \cite{wang2024variational}.
Finally, the output is measured, and the corresponding quantum states are decoded back into classical information by computing the expected values of the qubits with respect to a given observable, typically Pauli-Z. The resulting outcomes represent the predicted values.

The embedding layer utilized in the VQR model employs angle embedding \cite{lloyd2020quantum}, a method to encode classical data into quantum states by adjusting the angles of quantum gates in variational circuits.
This technique aligns the rotation angles of gates within the quantum circuit to classical data values, offering an efficient route for embedding classical information into quantum states.
Specifically, angle embedding manipulates a quantum state starting from $\ket{0}$ by applying rotations based on classical data through Pauli matrices ($X$, $Y$, $Z$) across different spatial axes.
For instance, applying a rotation $RY(\theta)$ gate to a qubit initially in the state $\ket{0}$ transforms it into a superposition state:
\begin{equation}
    \label{eq:angle-embedding-example}
    RY(\theta)\ket{0} = \cos(\frac{\theta}{2})\ket{0} + \sin(\frac{\theta}{2})\ket{1}.
\end{equation}
This results in quantum states that are uniquely determined by the classical inputs, setting the stage for subsequent quantum operations.

For the constructed VQR model, it was chosen to use the Strongly Entangling Layers from PennyLane as ansatz.
\begin{figure}[htbp]
    \centerline{\includegraphics[width=1.2\linewidth]{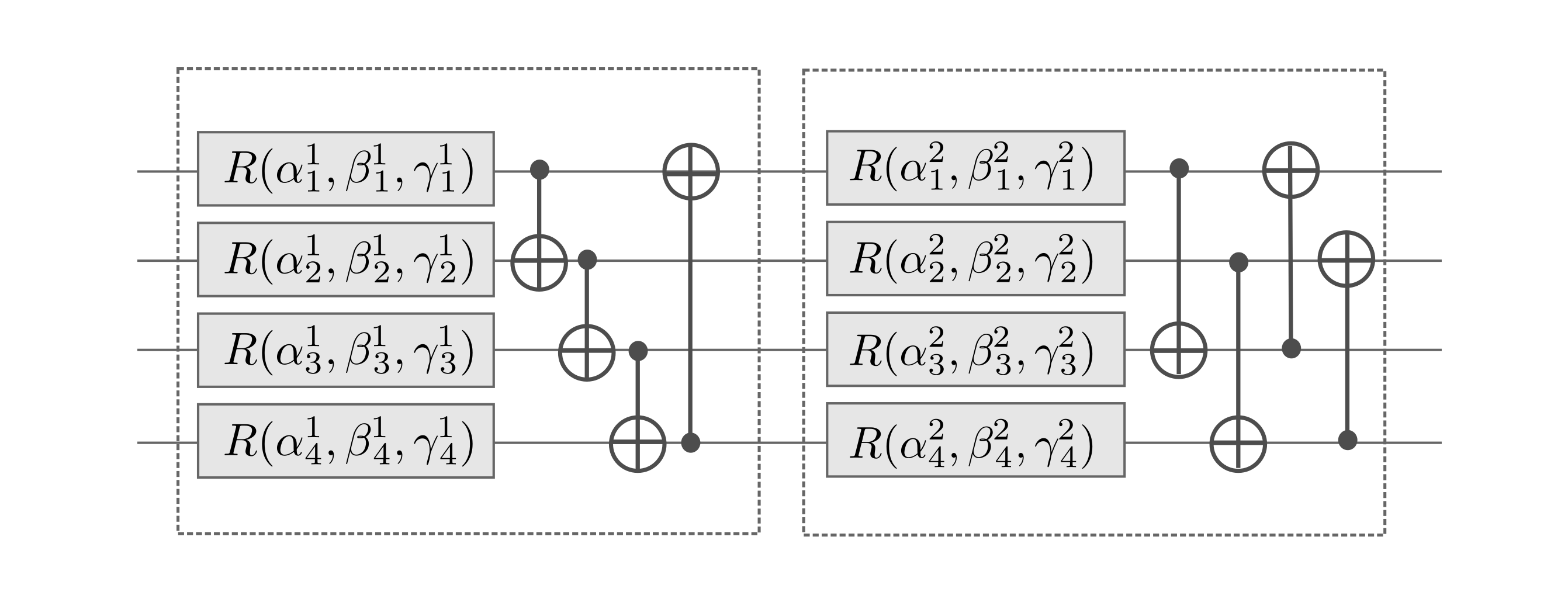}}
    \caption{The Strongly Entangling Layers with 2 layers, from Pennylane website.}
    \label{fig:strongly-entangl-layer}
\end{figure}
The repeated layers (Fig.~\ref{fig:strongly-entangl-layer}) that are applied to the qubits one after the other form the basis of this kind of ansatz. 
The purpose of each layer is to encourage significant entanglement between qubits.

Two types of circuits representing a VQR model were tested in this work:
\begin{itemize}
    \item \textbf{Linear:} This architecture is composed of a single Angle Embedding layer from PennyLane, followed by several Strongly Entangling layers representing the ansatz.
    \item \textbf{Non-linear:} Multiple Angle Embeddings are employed within the quantum circuit (Fig.~\ref{fig:vqr-non-linear}), allowing the quantum model to access a broader frequency spectrum and empowering it to identify and learn more complex patterns in the data \cite{PhysRevA.103.032430}. 
    This enables the realization of all Fourier coefficients, making these models capable of approximating any function and achieving a universal function approximation.
\end{itemize}

\subsection{Quantum Long Short-Term Memory (QLSTM)}
The QLSTM model, introduced in~\cite{chen2022quantum}, is a fusion of LSTM networks with quantum computing, with the aim of maximizing quantum efficiency in handling sequential data.
Similar to its classical counterpart, the QLSTM attempts to model temporal dependencies but does so within the framework of a quantum circuit.
Starting from the LSTM architecture, which is widely employed in time series forecasting due to its capability to capture long-term dependencies in sequential data, the key innovation behind the QLSTM lies in integrating a VQC layer within two classical Fully Connected layers, thereby encoding input data into quantum states using quantum feature maps and ansatzes.
Subsequently, by manipulating these states, quantum circuits that carry out calculations in a high-dimensional feature space may be able to reveal intricate patterns in the data that would otherwise remain hidden using conventional computations.

\begin{figure}[htbp]
    \centerline{\includegraphics[width=\linewidth]{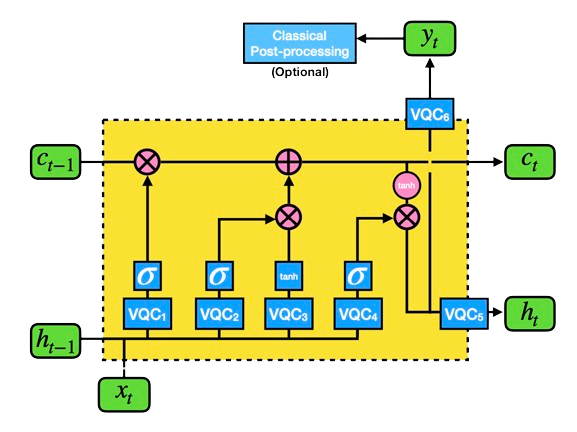}}
    \caption{The QLSTM architecture. From \cite{chen2022quantum}.}
    \label{fig:qlstm-architecture}
\end{figure}

In a classical LSTM, the operation at each time step involves several gates — the forget gate, input gate, and output gate — which control the flow of information through the cell. 
These gates determine what information is retained or discarded as the sequences are processed, allowing LSTMs to handle long-term dependencies within the data effectively.
\\
QLSTM replaces the classical components of the LSTM gates with quantum circuits (Fig.~\ref{fig:qlstm-architecture}). 
These VQCs are designed to perform similar gate operations but utilize quantum properties like entanglement and superposition, aiming to enhance the model's ability to capture complex patterns in data.

From Fig.~\ref{fig:qlstm-architecture} it is possible to observe the entire transition from the classic LSTM to its quantum version.
The $\sigma$ and $tanh$ blocks represent the sigmoid and the hyperbolic tangent activation function, respectively.
In addition, $x_t$ is the input at time $t$, $h_t$ is for the hidden state, $c_t$ is for the cell state, and $y_t$ is the output.
\\
The forget vector $f_t$ is computed using a VQC, denoted as $\mathrm{VQC}_1$:
\begin{equation}
    f_t = \sigma(\mathrm{VQC}_1(v_t)).
\end{equation}

Similarly, the input vector $i_t$ is computed using another VQC, $\mathrm{VQC}_2$:
\begin{equation}
    i_t = \sigma(\mathrm{VQC}_2(v_t)).
\end{equation}

The candidate memory state $\tilde{C}_t$ is generated by $\mathrm{VQC}_3$:
\begin{equation}
    \tilde{C}_t = \tanh(\mathrm{VQC}_3(v_t)).
\end{equation}

The cell state is updated using the outputs of the quantum gates:
\begin{equation}
    C_t = f_t * C_{t-1} + i_t * \tilde{C}_t.
\end{equation}

The output gate vector $o_t$ is computed using $\mathrm{VQC}_4$:
\begin{equation}
    o_t = \sigma(\mathrm{VQC}_4(v_t)).
\end{equation}

The hidden state and the final output are computed using additional VQCs, i.e. $\mathrm{VQC}_5$ and $\mathrm{VQC}_6$:
\begin{equation}
    h_t = \mathrm{VQC}_5(o_t * \tanh(C_t)),
\end{equation}
\begin{equation}
    y_t = \mathrm{VQC}_6(o_t * \tanh(C_t)).
\end{equation}

\begin{figure}[htbp]
    \centerline{\includegraphics[width=\linewidth]{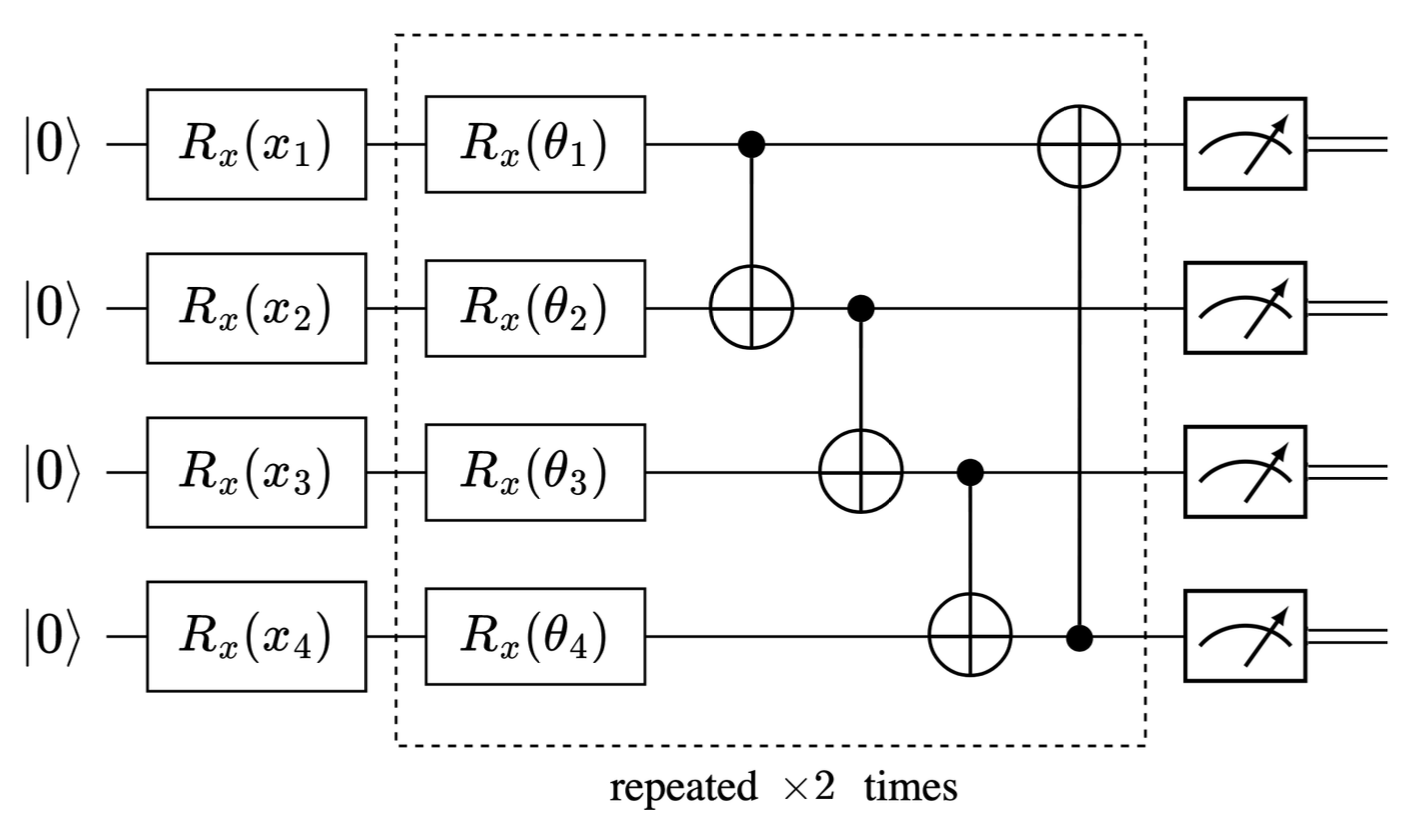}}
    \caption{Circuital schema of the proposed VQC inside the QLSTM model.}
    \label{fig:qlstm-circuit}
\end{figure}

As in the VQR case, the VQC in the QLSTM is composed of three main layers: the data encoding layer, the variational layer, and the quantum measurement layer. 
Each of these layers plays a specific role in processing the input data, manipulating quantum states, and extracting useful information for further classical processing.
Fig.\ref{fig:qlstm-circuit} shows the architecture of the VQC layer's design inside the QLSTM.

The VQC in Fig.\ref{fig:qlstm-circuit} begins with a layer of data encoding that applies $R_x$ gates.
In quantum computing, the $R_x$ gate is a rotation around the Bloch sphere's $x$-axis that modifies the qubits' state. 
The $R_x$ gates are used in VQCs to convert classical input data into quantum states. 
Since it converts real-valued data into a format that can be processed by a quantum circuit, this encoding is crucial and it allows to effectively project input data into a high-dimensional feature space in order to enhance information processing inside the circuit \cite{lloyd2020quantum}.
After encoding the data, the circuit employs the variational layer with an ansatz.
The ansatz is made of adjustable $R_x$ gates, optimized during training to tailor quantum operations to specific tasks.
This allows the circuit to adapt based on the input and required output.
Additionally, the design incorporates circular CNOT gates to entangle qubits in a ring-like structure, facilitating entanglement across the circuit and enabling complex state transformations by creating correlations within the quantum system.
The measurement of every qubit in the Pauli-Z basis is the last phase in the VQC.
The spin along the $z$-axis (up or down, corresponding to $1$ or $-1$) is determined by the Pauli-Z measurements.
The expected values, which are essential for the VQC output, are provided by these measurements.
The expected value of an observable in quantum mechanics provides the average result of several repetitions of the same experiment on identically prepared systems.

\section{Methodology}
\label{sec:methodology}
In addressing the pressing need for more accurate air quality monitoring, this research compares classical DL models and QML algorithms to calibrate low-price optical fine dust sensors.
The calibration process is in fact essential to ensure the quality of the data produced by these sensors.

\subsection{Dataset and Data Preprocessing}
The dataset~\cite{zenodo_dataset} was constructed during a measurement campaign that lasted seven months, from November 2022 to May 2023.
24 low-cost dust sensors measuring $\mathrm{PM}_{2.5}$ were positioned near the inlet of a reference-grade monitoring station of ARPA (the Italian Regional Environmental Protection Agency).
The reference station is located in a suburban area in the city of Turin.
Measurements from both the sensors and the official monitoring station were collected, the former having a sampling rate of one second while the latter of one hour.
Temperature, humidity, and atmospheric pressure were also sampled by nine low-cost sensors every 2-5 seconds, to characterize the environmental conditions.
%
%
The primary aim is to calibrate the readings from the low-cost sensors against the more accurate, reference-grade measurements, addressing discrepancies due to environmental factors and sensor inaccuracies.

Key pre-processing steps include the reduction of data-granularity, by taking both the minute and the hour averages of all the low-cost sensor measurements.
Subsequently, the median of all the low-cost sensors is computed to obtain a single time series for each monitored quantity in both temporal aggregations.
%
%

This is followed by the adaptation of the data for compatibility with quantum computing paradigms.
This includes encoding the processed data into a quantum-friendly format.
In fact, in order to utilize the VQR model, the input features are scaled in a range from $[-1, +1]$ since the $arctan$ function is used as the feature map of the embedding layer. Similarly, output features for our quantum models are also scaled in a $[-1, +1]$ range to match the range of Pauli-Z expected values. 
For the QLSTM model, there is no need to scale the input features, because it uses a simple angle embedding within a hybrid version of the classical LSTM model.

\subsection{Model Implementation}
Some preliminary tests were conducted to select the input features and their temporal aggregation for each model. In particular, an extensive grid search procedure and cross validation were carried out for each model and hyperparameter combination.

The best configuration of FFNN uses as input features the hour measurements of the low-cost sensors, i.e. $\mathrm{PM}_{2.5}$, temperature, humidity, and atmospheric pressure, targeting the reference hour values of $\mathrm{PM}_{2.5}$ of ARPA.
The same configuration is also selected for VQR, to allow for a better comparison between the two.

LSTM, instead, seems not to take advantage of the additional environmental parameters.
For this reason, only the hourly aggregated $\mathrm{PM}_{2.5}$ time series is selected as an input feature.
In order to act as a calibration function, the model is trained on a window of T samples while targeting the reference value of $\mathrm{PM}_{2.5}$ measured in the same hour as the last sample in the window.
This model structure was also selected for QLSTM.


In the implementation, training and testing of the models leverage Python 3.9, with key libraries including Pandas for data manipulation, PyTorch for building and training neural networks, and PennyLane for constructing quantum circuits. 
The codebase, available as open-source on Github~\cite{github}, is structured to facilitate the comparison of models through a unified framework, allowing for streamlined experimentation and analysis.

The research results were obtained by running the source code on a classical machine, simulating the behavior of a quantum one.

\subsection{Training and Testing Procedure}
The training and testing of the models are performed in different steps.
At first, a meticulous tuning process is undertaken for each model, involving a grid-search optimization of hyperparameters to achieve the best possible calibration accuracy.
This is performed considering a fixed partition of the training and test set.
Since LSTM and QLSTM need to train on a continuous time series, it was decided to perform a chronological split of the data into the train and test set, without shuffling.
The hyperparameter search space includes the number of epochs, learning rate, optimizer choice, and the specific configurations for each model, such as the number of layers and qubits for quantum models.
For what concerns the loss function, the L1 norm proves to be the most effective for all models except VQR, which instead benefits from using MSE.

The second step is to perform k-fold cross-validation for the optimized model parameters.
For FFNN and VQR, the dataset is randomized before splitting it into different folds.
For LSTM and QLSTM this is not possible, since the training and testing require continuous data.
However, non-contiguous folds still have to be joined to create the training datasets.

Finally, the results of the models' cross-validation are compared with each other and against a benchmark that estimates the improvement achieved by the calibration.
The benchmark is realized by applying the loss functions to non-calibrated data, by selecting multiple random sample sets of the same size of the folds. 


\section{Results}
\label{sec:results}
The comparative analysis of FFNN, LSTM, VQR, and QLSTM models yields a comprehensive understanding of their performance in calibrating low-cost optical fine dust sensors.
This comparison was carried out with pairs of models: FFNN with its quantum counterpart VQR, and LSTM with QLSTM.

\subsection{Hyperparameters Tuning}

For the classical FFNN and LSTM models, various hyperparameter configurations were tested, including different learning rates (ranging from 0.001 to 0.1), number of layers (from 1 to 3 for FFNN, and 1 to 2 for LSTM), and neuron counts in each layer. 
The optimal settings were found to be a learning rate of $0.0001$, three hidden layers, and 30-15-5 neurons per layer for the FFNN, and a learning rate of $0.001$, two recurrent layers, and 15 as the number of features in the hidden state $h$ (i.e. the hidden size) for the LSTM model.

The quantum models, VQR and QLSTM, required tuning of quantum-specific parameters, such as the number of qubits and variational circuit depth.
The VQR model achieved its best performance with 4 qubits and 4 layers of quantum circuits, while the QLSTM model performed optimally with 5 qubits and a depth of 7 layers in its variational circuits.

Let us compare the results of this step by considering classical models against their quantum counterparts.
Table~\ref{tab:ffnn-vqr-hyperparams-comparison} shows the results of FFNN and VQR, and Table~\ref{tab:lstm-qlstm-hyperparams-comparison} those obtained by hyperparameters tuning of LSTM and QLSTM.
\begin{table}[htbp]
    \caption{Best hyperparameter configuration obtained through the tuning phase from the FFNN and VQR models.}\label{tab:ffnn-vqr-hyperparams-comparison}
    \begin{center}
    \begin{tabular}{l|ll}
    \toprule
    \textbf{Hyperparameter} & \textbf{FFNN} & \textbf{VQR} \\
    \midrule
    Training-size & 75\% & 75\% \\
    N° epochs & 200 & 200 \\
    Learning rate & 0.0001 & 0.01 \\
    Optimizer & SGD & Adam \\
    Criterion & L1 & MSE \\
    Batch size & 10 & 10 \\
    Hidden size l1 & 30 & 4 \\
    Hidden size l2 & 15 & - \\
    Hidden size l3 & 5 & - \\
    N° qubits & - & 4 \\
    Type of architecture & - & Linear \\
    \midrule
    \textbf{Loss on training-set} & 6.790 & $8.699$ $(6.274)^*$ \\
    \textbf{Loss on test set} & 2.920 & $5.794$ $(4.806)^*$  \\
    \bottomrule
    \end{tabular}\\
    \end{center}
    \vspace{3pt}
    ($^*$) The VQR losses are reported as RMSE (instead of MSE) to be expressed as $\mathrm{PM}_{2.5}$ concentrations ($\mu g/m^3$). L1 norm is also shown (between round brackets) for VQR, to allow for a comparison.
\end{table}
%
%
The performance results from the FFNN and VQR models (Table~\ref{tab:ffnn-vqr-hyperparams-comparison}) are obtained using as loss functions the L1 norm and MSE, respectively.
For VQR, results are reported as RMSE to be expressed as $\mathrm{PM}_{2.5}$ concentrations.
To allow for a comparison, the VQR L1 losses computed on the training and test set are also shown.
While on the training set the VQR loss is slightly lower, the FFNN performs better on the test set.
%
%
%
\begin{table}[htbp]
    \centering
      \caption{Best hyperparameter configuration obtained through the tuning phase from the LSTM and QLSTM models.}
    \begin{tabular}{l|ll}
    \toprule
    \textbf{Hyperparameter} & \textbf{LSTM} & \textbf{QLSTM} \\
    \midrule
    Training-size & 70\% & 70\% \\
    N° epochs & 300 & 400 \\
    Learning rate & 0.001 & 0.01 \\
    Optimizer & RMSprop & Adam \\
    Criterion & L1 & L1 \\
    T (hours used to calibrate) & 3 & 5 \\
    Hidden size l1 & 15 & 15 \\
    N° recurrent layers & 2 & - \\
    N° quantum layers & - & 7 \\
    N° qubits & - & 5 \\
    Type of ansatz & - & Strongly \\
    \midrule
    \textbf{Loss on training-set} & 5.029 & 5.202 \\
    \textbf{Loss on test set} & 2.768 & 2.698  \\
    \bottomrule
    \end{tabular}
    \label{tab:lstm-qlstm-hyperparams-comparison}
\end{table}
The results for LSTM and QLSTM are compared in Table~\ref{tab:lstm-qlstm-hyperparams-comparison}.
It can be seen how the QLSTM model has a slightly lower error rate than LSTM.
It is important to note that for LSTM and QLSTM the configuration search space of hyperparameters, in terms of hidden size and number of layers, was chosen to be similar between the two models.
This makes for a fairer comparison since bigger LSTM models can provide better results, but QLSTM models of greater complexity cannot be simulated on classical hardware.
In any case, the optimized LSTM and QLSTM models respectively use a total of $482$ and $66$ trainable weights, resulting in a lower complexity for the QLSTM model.
%
%
Conversely, extending the hyperparameter search space of FFNN does not lead to better solutions.


Both tables show that the models perform better on the test set w.r.t. the training set.
This is a consequence of using a chronological partition of the training and test sets.
Due to the seasonality of $\mathrm{PM}_{2.5}$ in the city of Turin, late spring concentrations, which correspond to the test set, are lower and therefore affected by smaller measurement errors.
This limitation is addressed in the cross-validation phase.

Table~\ref{tab:loss-benchmark}, shows the loss functions applied to the original and non-calibrated data of the test sets.
The comparison of these values with the results obtained by the trained models, confirms that all of them achieve an improvement in terms of data quality (VQR only with RMSE).

\begin{table}[htbp]
    \centering
     \caption{Benchmark results for each loss function on the test set.}
    \begin{tabular}{l|lll}
    \toprule
    Benchmarks             & \textbf{L1} & \textbf{MSE} & \textbf{RMSE} \\ \midrule
    \textbf{Test set 30\%} & 5.030           & 42.533            & 6.522             \\
    \textbf{Test set 25\%} & 4.701           & 33.914            & 5.823             \\
    \bottomrule
    \end{tabular}
    \label{tab:loss-benchmark}
\end{table}

\subsection{Cross-validation Results}
%
Determining how different sections of the dataset affect the training and testing of the models, including the potential for overfitting, is crucial.
Cross-validation is a common method in ML to assess how well a model is able to generalize.
%
%
K-fold cross-validation is a common approach where the dataset is divided into 'k' segments.
The model trains on 'k-1' segments and is tested on the remaining one, repeating this process so that each segment serves as the test set once.
By averaging the results from all 'k' rounds, this technique offers a comprehensive evaluation, ensuring that every data point contributes to both training and testing, maximizing data utilization.
Tables~\ref{tab:cv-results-ffnn-vqr} and~\ref{tab:cv-results} show the cross-validation results for each model. 
The number of folds used varies according to the model: for the validation of FFNN and VQR, $K=4$ was chosen, while for the other two models, $K=5$ was chosen.
The reason for this difference is that for FFNN and VQR it was decided to maintain the same training and test set sizes of the hyperparameter tuning phase.
For these models, the data was shuffled before selecting the partitions.
LSTM and QLSTM, instead, require the data to be continuous during training and testing.
Therefore, it was decided to use more folds, to better comprehend the effect of seasonality.


\begin{table}[htbp]
    \centering
    \caption{Cross-validation result of FFNN and VQR model.}
    \begin{tabular}{l|llll|l}
    \toprule
    Cross-validation & \textbf{Fold-1} & \textbf{Fold-2} & \textbf{Fold-3} & \textbf{Fold-4} & \textbf{Avg} \\
    \midrule
    \textbf{FFNN} & 5.858 & 5.727 & 5.466 & 10.889 & 6.985 \\
    \textbf{VQR (L1)} & 11.249 & 10.916 & 9.730 & 10.073 & 10.492 \\
    \textbf{VQR (RMSE)} & 13.720 & 13.563 & 11.956 & 13.089 & 13.082 \\
    \bottomrule
    \end{tabular}
    \label{tab:cv-results-ffnn-vqr}
\end{table}

\begin{figure}[htbp]
    \centerline{\includegraphics[width=\linewidth]{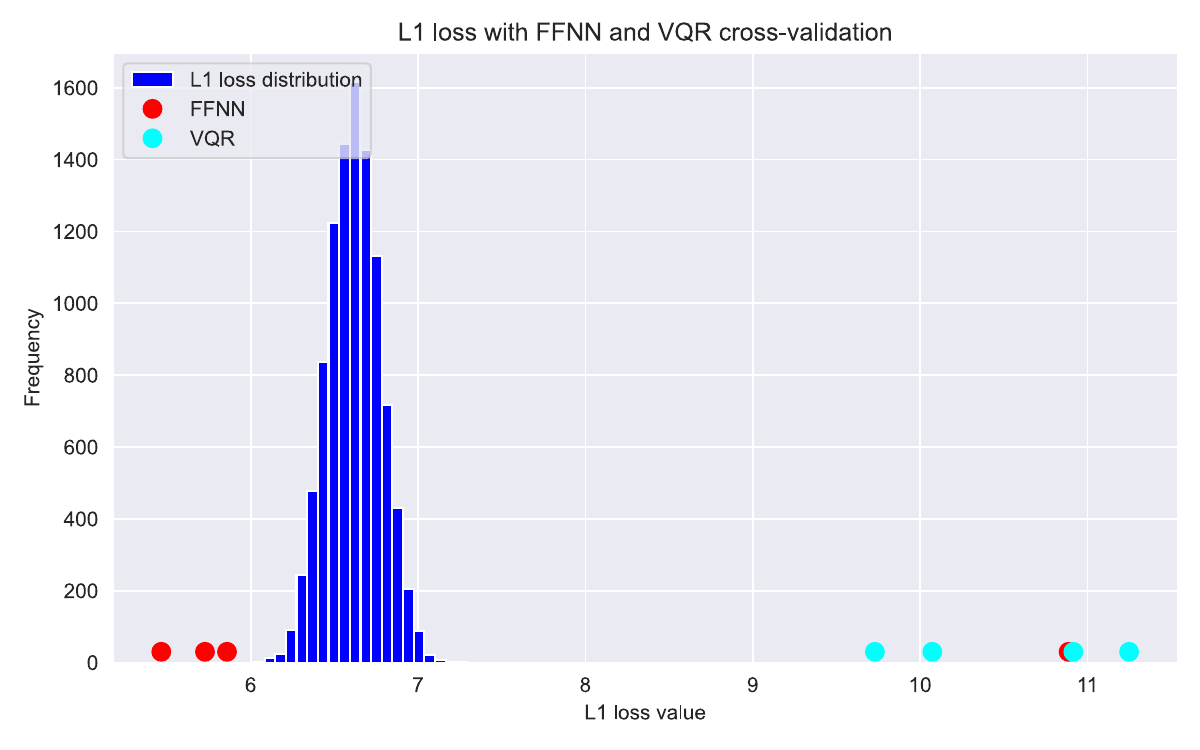}}
    \caption{The cross-validation result of the FFNN and VQR models
compared to their benchmark.}
    \label{fig:ffnn-vqr-cv-result}
\end{figure}

Table~\ref{tab:cv-results-ffnn-vqr} presents the results of cross-validation for the FFNN and VQR models.
VQR results are also evaluated with the L1 norm to allow a comparison with FFNN.
Cross-validation confirms the superiority of the FFNN model across all folds.
Fig.~\ref{fig:ffnn-vqr-cv-result} plots the result of the models against the benchmark distribution.
This is generated by computing the L1 loss on randomly selected sample sets of non-calibrated data with the same size of the folds.
From the image, it can be seen that the FFNN model is able to improve the quality of the non-calibrated data in most circumstances, while the VQR achieves the opposite effect.

\begin{table}[htbp]
    \centering
    \caption{Cross-validation results of LSTM and QLSTM models.}
    \begin{tabular}{l|lllll|l}
    \toprule
    Cross-valid. & \textbf{Fold-1} & \textbf{Fold-2} & \textbf{Fold-3} & \textbf{Fold-4} & \textbf{Fold-5} & \textbf{Avg} \\
    \midrule
    \textbf{LSTM} & 4.44 & 7.26 & 5.57 & 3.60 & 2.50 & $4.67$ \\
    \textbf{QLSTM} & 4.31 & 7.73 & 5.46 & 3.70 & 2.36 & $4.71$ \\
    \bottomrule
    \end{tabular}
    \label{tab:cv-results}
\end{table}

\begin{figure}[htbp]
    \centerline{\includegraphics[width=\linewidth]{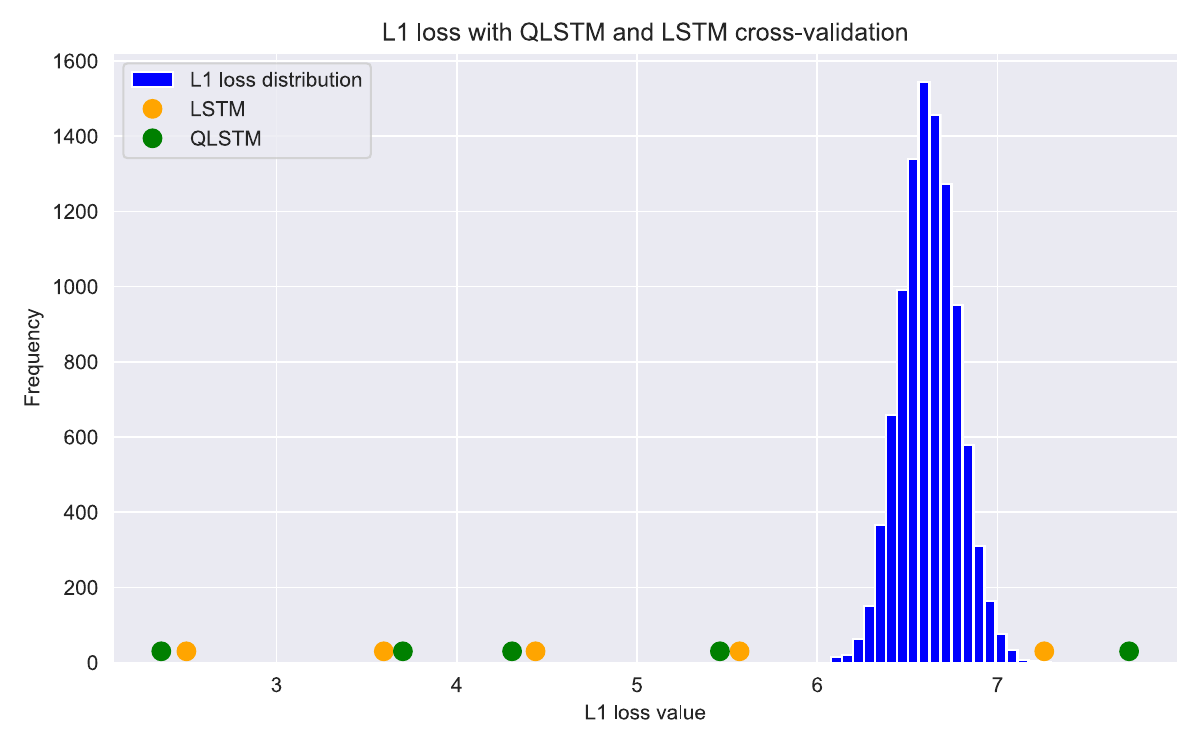}}
    \caption{The cross-validation result of the LSTM and QLSTM models
compared to their benchmark.}
    \label{fig:lstm-qlstm-cv-result}
\end{figure}

Table~\ref{tab:cv-results} presents the results of LSTM and QLSTM.
Averaging the result of the loss function in each fold shows that the LSTM model has an average of $4.67$ compared to the QLSTM model which gets to $4.71$.
The two models perform very similarly across the folds, with the quantum model managing to achieve slightly better results in some of them.
This is a remarkable result for the QLSTM model, since it achieves similar performance with far fewer trainable weights compared to LSTM: $66$ against $482$.
The seasonality of the $\mathrm{PM}_{2.5}$ concentrations is also evident in the table, where the test folds are presented following the chronological order of the contained data.
In fact, the loss increases during winter and decreases when reaching the end of spring.
Fig.~\ref{fig:lstm-qlstm-cv-result} plots the cross-validation results of LSTM and QLSTM against the benchmark distribution, showing that both models achieve significant improvements in terms of data quality in most cases.

\subsection{Graphical Comparison}
\label{sec:comparison}

\begin{figure}[htbp]    \centerline{\includegraphics[width=\linewidth]{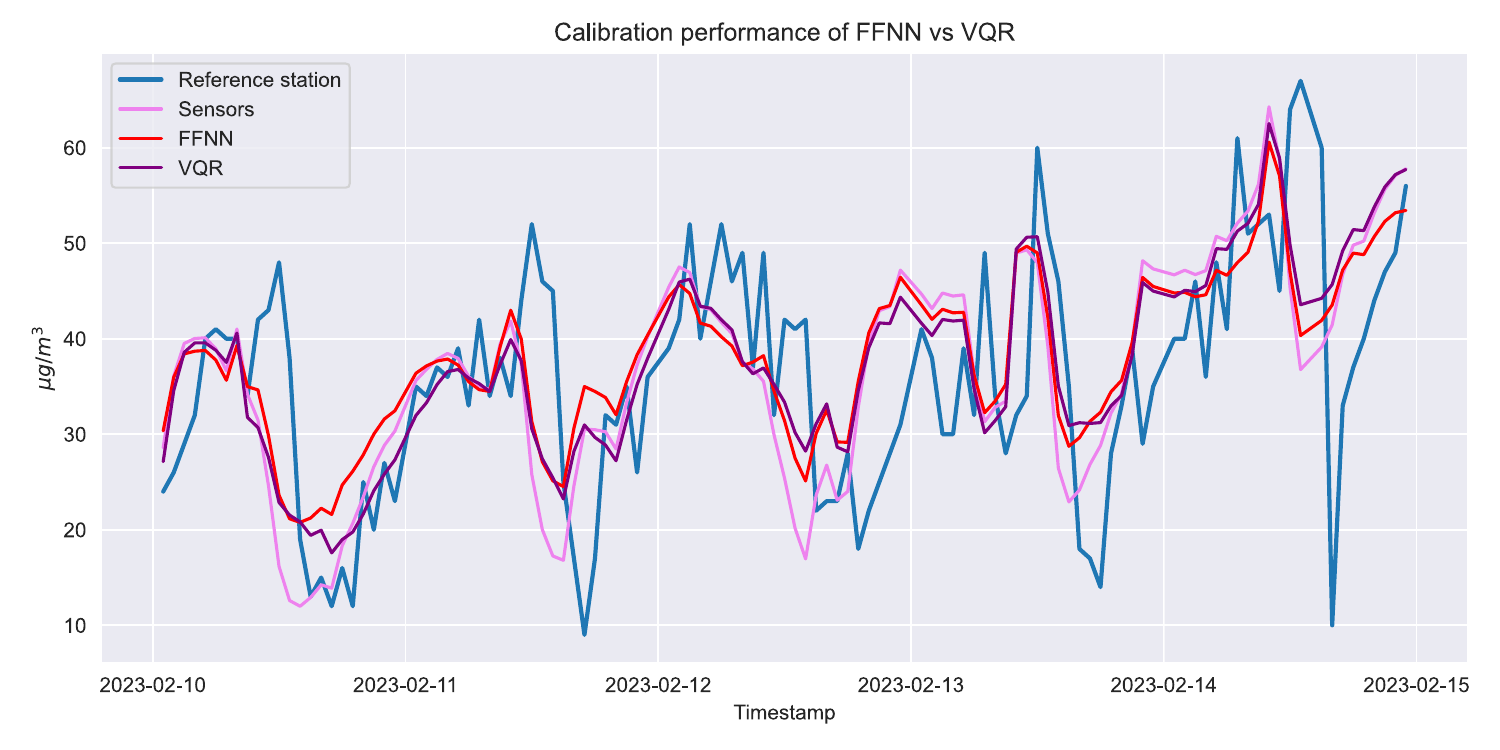}}
    \caption{The calibration performance obtained from FFNN and VQR models in comparison with the reference station values, observing a limited time range from February 10 to 15, 2023.}
    \label{fig:ffnn-vqr-models-performance}
\end{figure}

\begin{figure}[htbp]
    \centerline{\includegraphics[width=\linewidth]{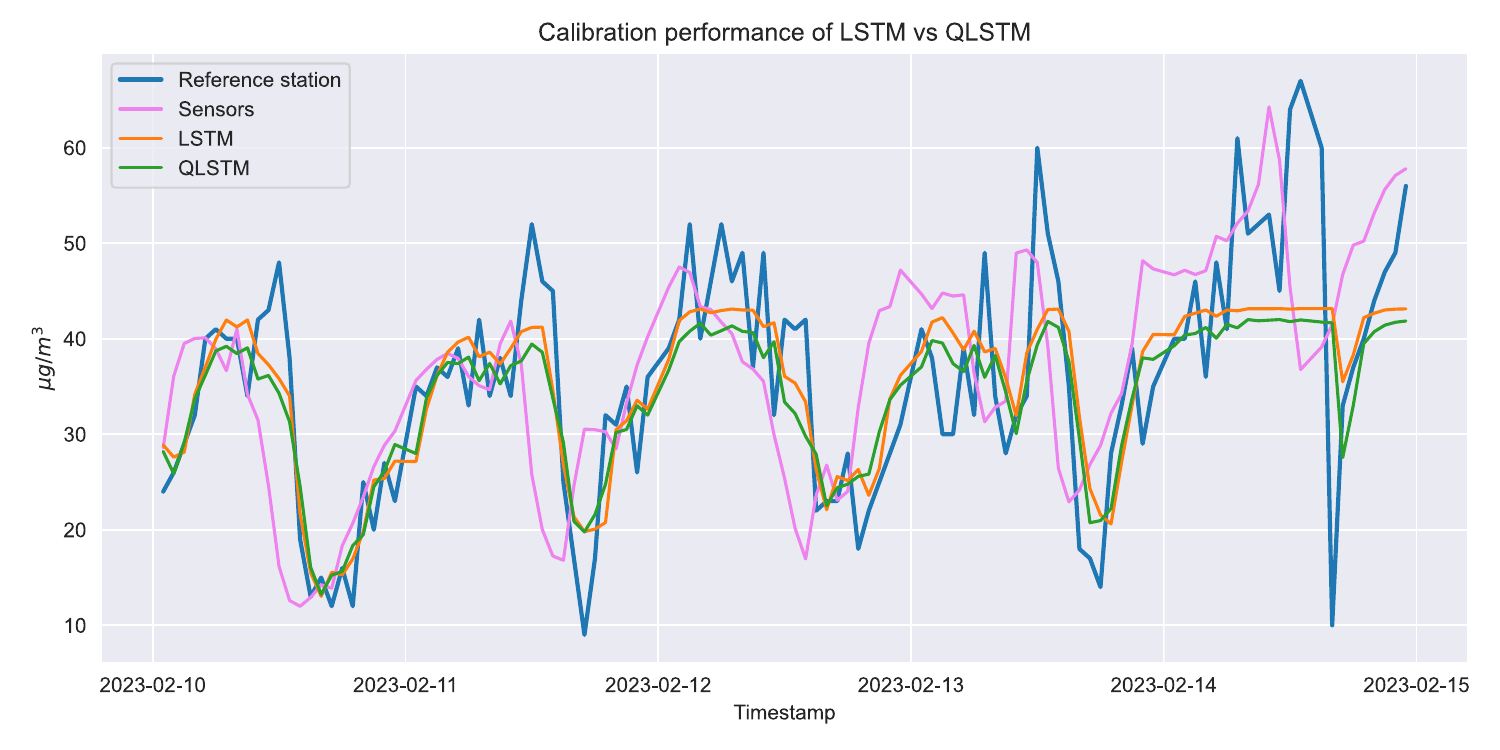}}
    \caption{The calibration performance obtained from LSTM and QLSTM models in comparison with the reference station values, observing a limited time range from February 10 to 15, 2023.}
    \label{fig:lstm-qlstm-models-performance}
\end{figure}

Figures~\ref{fig:ffnn-vqr-models-performance} and~\ref{fig:lstm-qlstm-models-performance} plot the non-calibrated and ARPA reference data against the output of the different calibration models, over a window of five days.
This provides a better visual representation of the effects of the different calibration models.

An interesting observation for LSTM and QLSTM (in Fig.~\ref{fig:lstm-qlstm-models-performance}) is that the calibration data tends not to follow very closely the local peaks of $\mathrm{PM}_{2.5}$ concentrations.
This may be attributed to the choice of L1 loss function during training, which does not penalize this behavior as much as RMSE or MSE would.

\section{Conclusion}
\label{sec:conclusions}
This research explores the application of QML to the calibration of low-cost optical fine dust sensors, an emerging technology for air quality monitoring within smart cities.
By comparing traditional DL models, i.e. FFNN and LSTM, with their quantum counterparts, that is VQR and QLSTM, this work showcases the potential of QML to become a valid tool in environmental science.

The LSTM and QLSTM models, with their ability to process temporal sequences, have emerged as particularly effective, confirming the importance of considering temporal patterns in environmental data.
In particular, QLSTM slightly outperforms LSTM, while utilizing less training parameters: 66 against 482.
The VQR model, instead, proves not to be as effective as its classical counterpart (FFNN).

Future works should explore bigger and more varied datasets, and further extend the hyperparameter search space to include more complex classical and quantum models.
Finally, testing on real quantum hardware should be performed to evaluate the consistency of the results and related computational advantages.

\section*{Acknowledgment}

This work has been partially supported by the ``NATIONAL CENTRE FOR HPC, BIG DATA AND QUANTUM COMPUTING'' (CN1) within the Italian ``Piano Nazionale di Ripresa e Resilienza (PNRR)'', Mission 4 Component 2 Investment 1.4 funded by the European Union - {NextGenerationEU} - CN00000013: A. Ceschini, M. Panella and A. Rosato in Spoke 10, CUP B83C22002940006, while Edoardo Giusto in Spoke 9, CUP E63C22000980007.

\bibliographystyle{IEEEtran}
\bibliography{ARTICLE}

\end{document}